\documentclass[11pt,a4paper]{article}
\PassOptionsToPackage{hyphens}{url}
\usepackage[utf8]{inputenc}
\usepackage[T1]{fontenc}
\usepackage{times}
\usepackage[margin=1in]{geometry}
\usepackage{amsmath,amssymb}
\usepackage{booktabs}
\usepackage{graphicx}
\usepackage{url}
\usepackage{hyperref}
\usepackage{natbib}
\usepackage{xcolor}
\usepackage{tabularx}
\usepackage{enumitem}

\hypersetup{colorlinks=true,linkcolor=blue,citecolor=blue,urlcolor=blue}

\title{Concurrent Criterion Validation of a Validity Screen\\for LLM Confidence Signals via Selective Prediction}

\author{Jon-Paul Cacioli\\
Independent Researcher, Melbourne, Australia\\
ORCID: 0009-0000-7054-2014\\
\url{https://github.com/synthiumjp/validity-scaling-llm}}

\date{}

\begin{document}
\sloppy
\maketitle

\begin{abstract}

Cacioli (2026d, 2026e) introduced a validity screening protocol for LLM confidence data. The protocol classifies models as Invalid, Indeterminate, or Valid based on clinical psychometric indices. Its limitations section flagged the absence of a deployment-facing criterion. Tier classifications had not been calibrated against downstream outcomes.

This paper provides a first concurrent criterion validation against selective prediction. The validation uses the same 20-model, 524-item dataset as the derivation study. It tests deployment-relevant separation between tiers but does not establish generalisation to new models or benchmarks.

Valid models (n = 14) show mean Type 2 AUROC = .624, 95\% bootstrap CI [.604, .647]. Invalid models (n = 3) show mean AUROC = .357 [.031, .522]. The group difference is large (d = 2.81, p = .002) and survives removal of the most extreme case. The tiers order monotonically. Invalid (.357) < Indeterminate (.554) < Valid (.624). Bootstrap P(monotonic) = .783.

Split-half cross-validation, screening on one half and testing AUROC on the other, yields median d = 1.77 (95\% CI [0.95, 3.00]) across 1,000 splits. P(d > 0) = 1.0. DeepSeek-R1, classified Invalid by massive inversion, drops from 85.3\% accuracy at full coverage to 11.3\% at 10\% coverage. Per-track item sensitivity predicts per-track AUROC ($\rho$ = .788, p < .001, n = 107 model-track observations).

The three-tier classification accounts for 47.0\% of the variance in AUROC ($\eta$$^2$ = .470). The screen predicts whether confidence-based selective prediction works. It does not yet establish that predictions transport to other benchmarks, elicitation formats, or model generations.

\end{abstract}

\section{Introduction}

\subsection{The gap}

Two companion papers established a validity screening framework for LLM confidence data. Cacioli (2026d) derived six validity indices mapped to PAI and MMPI-3 scales. Cacioli (2026e) extracted a portable protocol. Three core indices (L, Fp, RBS), one structural indicator (TRIN), one diagnostic statistic (r(confidence, correct)). Models are classified as Invalid, Indeterminate, or Valid.

The screener paper's limitations section stated that tier thresholds had not been calibrated against deployment outcomes. This paper provides that calibration on the same sample. If the screen works, Invalid models should fail at selective prediction. Valid models should succeed. The relationship should be strong, monotonic, and robust.

\subsection{Why selective prediction}

Selective prediction is the most direct test of whether confidence carries item-level information about correctness. A model that selectively presents its high-confidence items, and achieves higher accuracy by doing so, has a confidence signal that discriminates correct from incorrect responses. In signal detection theory terms (Cacioli, 2026a), this is Type 2 discrimination.

Selective prediction is used in deployment for abstention (Wen et al., 2025), routing, and safety-critical escalation. If a model's confidence is uninformative and a deployment system uses it for abstention, the system does not abstain on hard items. It abstains randomly. Or it abstains on the items the model got right. Phillips et al. (2026) show that entropy-based uncertainty is insufficient for safe selective prediction in LLMs, reinforcing the need for validity-aware screening.

The screen and the criterion are conceptually proximal. Both ask whether confidence discriminates correctness. The validation is therefore a near-neighbour test. It does not demonstrate that the screen predicts calibration error under distribution shift, human-AI collaboration outcomes, or cost-sensitive routing performance. It demonstrates that the screen predicts the operational behaviour most directly downstream of the construct it measures.

\subsection{What this paper contributes}

This is a concurrent criterion validation. The screen produces a classification. Selective prediction metrics produce continuous outcomes. We test whether the classification predicts the outcome.

The contribution is a construct-validity bridge between psychometric validity indices and ML deployment metrics. The screen was developed from clinical psychometric principles (PAI/MMPI-3 validity scaling). The criterion comes from the selective prediction literature (Geifman \& El-Yaniv, 2017; Feng et al., 2024). Showing that the former predicts the latter establishes that the clinical methodology transfers, not merely that the indices have face validity.

\subsection{Scope}

Selective prediction is one deployment-facing criterion, not the only one. Wang et al. (2026) show that verbalized confidence can be separable from actual calibration signals. Dai (2026) shows that probe scale design changes measured metacognitive efficiency. Wei et al. (2026) distinguish response calibration from capability calibration. The present study validates the screener against one operational endpoint. It does not validate confidence quality in every sense.

\section{Method}

\subsection{Models}

Twenty frontier LLMs from seven families. Anthropic (Claude Haiku 4.5, Sonnet 4.6, Opus 4.6). OpenAI (GPT-5.4, GPT-5.4 mini, GPT-5.4 nano). Google Gemini (2.5 Flash, 2.5 Pro, 3 Flash, 3.1 Pro). Google Gemma (1B, 12B, 27B). Qwen (235B, 80B Instruct, 80B Think, Coder 480B). DeepSeek (V3.2, R1). Zhipu (GLM-5).

The models are not independent samples from a single population. They cluster by family, architecture, and alignment regime. Family-level clustering is addressed in Section 3.6.

All models were evaluated on the Classical Minds metacognitive battery (Cacioli, 2026c). Each model produces item-level correctness labels and binary confidence probes (KEEP/WITHDRAW and BET/NO BET) across six cognitive tracks.

\subsection{Validity classifications}

The analysis pipeline calls \texttt{validity\_screen.py} (Cacioli, 2026e) directly on the aggregate KEEP/WITHDRAW data. Tier assignments are computed by the screen, not hard-coded.

Three models are classified Invalid. DeepSeek-R1 (Fp = .946, RBS = +.868, r = $-$.798). Gemini 3.1 Pro (L = .967, r = $-$.001). Qwen 80B Think (L = .974, r = +.047 n.s.). Gemini 3.1 Pro and Qwen 80B Think trigger the cell-count warning (min cell < 5) because they WITHDRAW very few items that they also get wrong. Their L values exceed the Invalid threshold and their r values are near zero. We retain the Invalid classification, noting the cell-count instability.

Three models are classified Indeterminate. GPT-5.4 nano (Fp = .369, r = +.119). Gemma 3 12B (Fp = .258, r = +.191). Gemma 3 1B (RBS = +.028, r = $-$.031). Gemma 3 1B was classified Invalid in Cacioli (2026e) with the note "marginally inverted, borderline." The screen classifies it as Indeterminate because the RBS confidence interval [$-$0.050, +0.107] includes zero. We use the screen's classification.

Fourteen models are classified Valid. All show positive, significant r(confidence, correct), ranging from +.071 to +.268.

\subsection{Data}

524 items per model across six cognitive tracks, following the task numbering in Cacioli (2026c). Learning/overhypothesis (T1, 98 items). Metacognition (T2, 90). Social cognition (T3, 116). Attention (T4, 60). Executive function (T5, 88). Prospective regulation (T6, 72). Total observations: 10,480.

Each item produces a correctness label, a KEEP/WITHDRAW response, and a BET/NO BET response. We construct an ordinal confidence score following Cacioli (2026d). KEEP+BET = 3, KEEP+NO BET = 2, WITHDRAW+BET = 1, WITHDRAW+NO BET = 0. AUROC is a rank-based measure and is robust to monotone transformations of the predictor. The ordinal encoding adds one bit of resolution over binary KEEP/WITHDRAW.

\subsection{Selective prediction metrics}

\textbf{Type 2 AUROC.} AUROC of ordinal confidence predicting binary correctness. 0.5 = chance. Below 0.5 = inverted. This is a non-parametric Type 2 discrimination measure (Fleming \& Lau, 2014).

\textbf{Selective gain.} At coverage c, items are sorted by confidence descending and the top c fraction retained. Selective gain = accuracy of retained set minus baseline accuracy.

\textbf{Risk-coverage curves.} Accuracy as a function of coverage from 100\% to 10\%.

\textbf{Risk-coverage AUC.} Trapezoidal integration of accuracy over coverage from 0.1 to 1.0. Higher = better. Some literature defines AURC as area under the risk curve, minimised. Our orientation (accuracy, maximised) is stated explicitly.

\subsection{Statistical tests}

Group differences. One-way ANOVA and Mann-Whitney U (Valid vs Invalid, one-tailed). Effect sizes. Cohen's d (pooled SD) and $\eta$$^2$. Bootstrapped 95\% CIs on tier means (10,000 samples). Within-tier correlations. Spearman's $\rho$. Split-half cross-validation. 1,000 random 50/50 item splits, screening on one half, AUROC on the other.

\section{Results}

\subsection{Type 2 AUROC by tier}

\begin{figure}[htbp!]
\centering
\includegraphics[width=0.95\columnwidth]{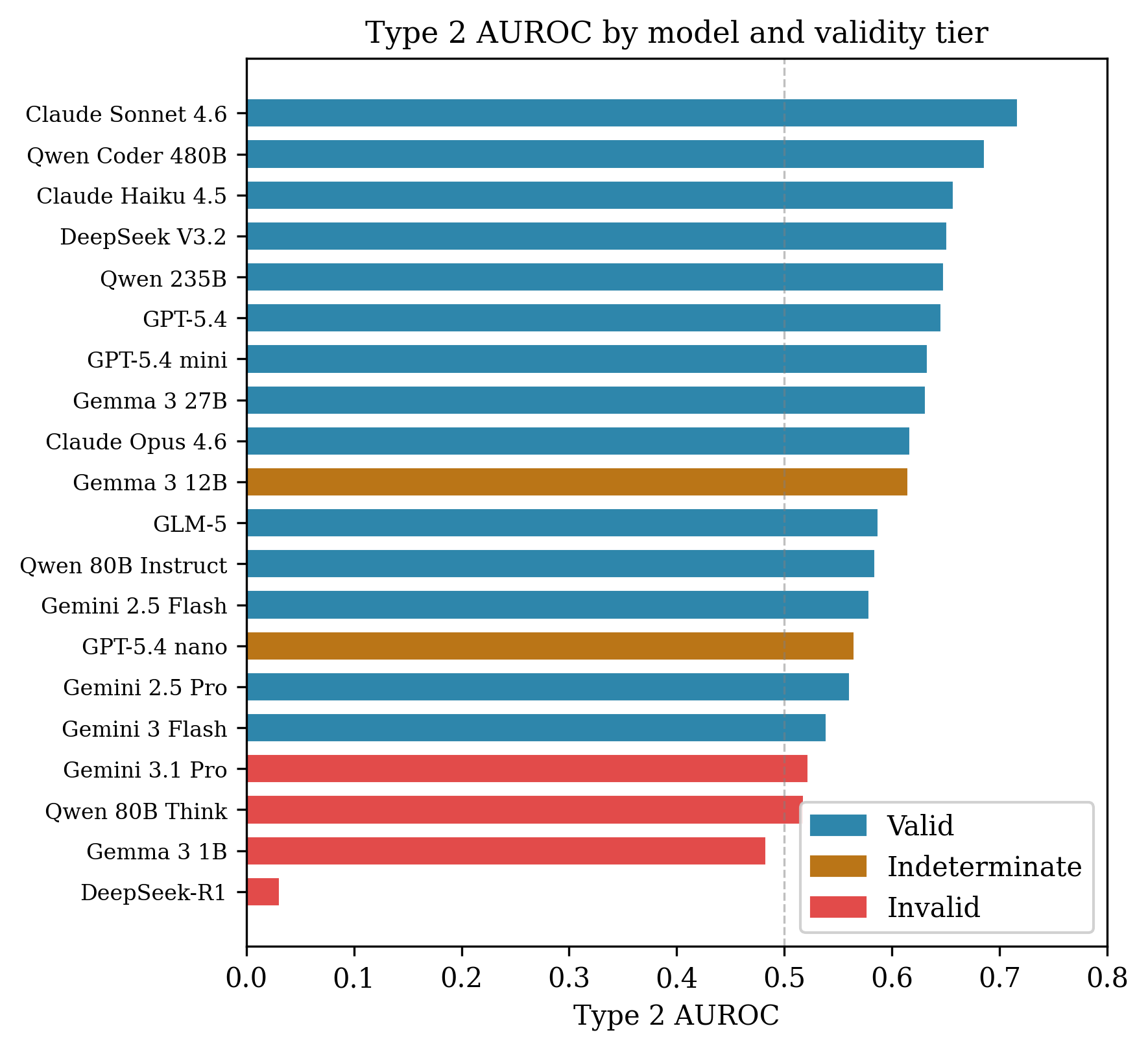}
\caption{Type 2 AUROC by model and validity tier.}
\label{fig:auroc}
\end{figure}

Table 1 reports AUROC, baseline accuracy, and selective gain for all 20 models.

\textbf{Table 1.} Type 2 AUROC and selective prediction gain by model and validity tier.

\begin{table}[htbp]
\centering\footnotesize
\begin{tabular}{lllllll}
\toprule
Model & Tier & Baseline & AUROC & $\Delta$80\% & $\Delta$70\% & $\Delta$50\% \\
\midrule
Sonnet 4.6 & Valid & .927 & .717 & +.039 & +.040 & +.050 \\
Qwen Coder 480B & Valid & .908 & .686 & +.032 & +.043 & +.042 \\
Claude Haiku 4.5 & Valid & .891 & .657 & +.040 & +.041 & +.044 \\
DeepSeek V3.2 & Valid & .885 & .651 & +.015 & +.030 & +.057 \\
Qwen 235B & Valid & .903 & .648 & +.035 & +.040 & +.029 \\
GPT-5.4 & Valid & .924 & .646 & +.033 & +.038 & +.046 \\
GPT-5.4 mini & Valid & .884 & .633 & +.021 & +.040 & +.048 \\
Gemma 3 27B & Valid & .880 & .631 & +.030 & +.041 & +.040 \\
Opus 4.6 & Valid & .937 & .617 & +.018 & +.022 & +.032 \\
GLM-5 & Valid & .943 & .587 & +.024 & +.022 & +.023 \\
Qwen 80B Inst & Valid & .895 & .584 & +.024 & +.029 & +.040 \\
Gemini 2.5 Flash & Valid & .939 & .579 & +.016 & +.026 & +.034 \\
Gemini 2.5 Pro & Valid & .926 & .561 & +.024 & +.020 & +.017 \\
Gemini 3 Flash & Valid & .950 & .539 & +.004 & +.003 & $-$.004 \\
Gemma 3 12B & Indet. & .857 & .615 & +.029 & +.045 & +.044 \\
GPT-5.4 nano & Indet. & .796 & .565 & +.030 & +.049 & +.044 \\
Gemma 3 1B & Indet. & .603 & .483 & $-$.003 & $-$.009 & +.126 \\
Gemini 3.1 Pro & Invalid & .943 & .522 & +.007 & +.014 & +.000 \\
Qwen 80B Think & Invalid & .926 & .518 & +.024 & +.023 & +.036 \\
DeepSeek-R1 & Invalid & .853 & .031 & $-$.036 & $-$.060 & $-$.143 \\
\bottomrule
\end{tabular}
\end{table}

\textit{Gain = accuracy at coverage threshold minus baseline. AUROC on ordinal confidence (0-3). N = 524 per model.}

Valid models show mean AUROC = .624 (SD = .048, bootstrap 95\% CI [.604, .647]). All 14 exceed chance. Invalid models show mean AUROC = .357 (SD = .231, CI [.031, .522]). Indeterminate models show mean AUROC = .554 (SD = .054).

Cohen's d = 2.81. Mann-Whitney U = 42, p = .002. Without DeepSeek-R1, Invalid mean AUROC = .520, U = 28, p = .008.

The tiers order monotonically. Invalid (.357) < Indeterminate (.554) < Valid (.624). Bootstrap P(monotonic) = .783.

\subsection{Selective gain by tier}

\begin{figure}[htbp!]
\centering
\includegraphics[width=0.7\columnwidth]{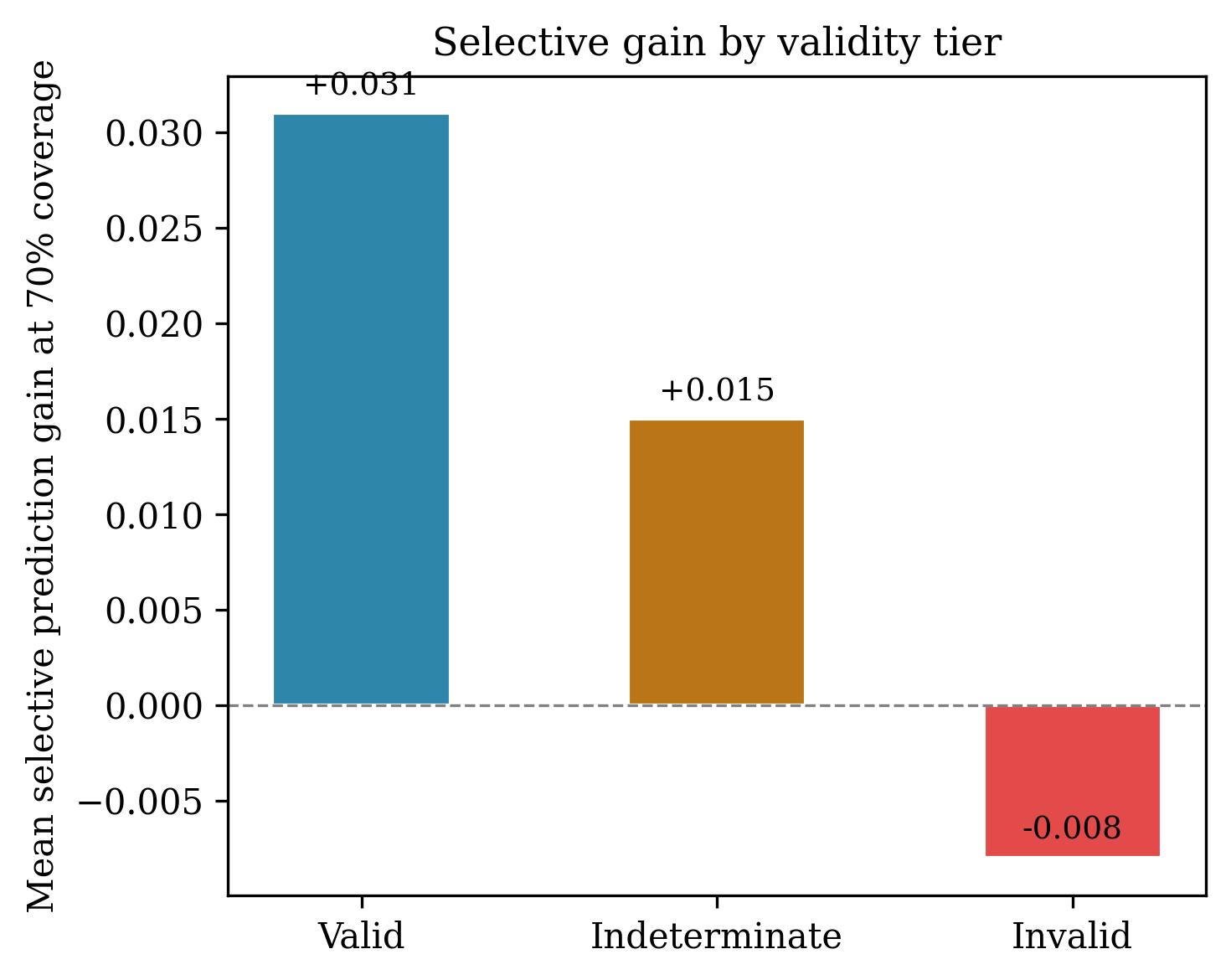}
\caption{Selective prediction gain at 70\% coverage by validity tier.}
\label{fig:gain70}
\end{figure}

At 70\% coverage, tier classification predicts selective gain. F(2,17) = 3.99, p = .038. Valid models gain 3.1 percentage points over baseline. Invalid models lose 0.8. Mann-Whitney U = 37, p = .025.

At 50\% coverage, the ANOVA is non-significant (p = .508). This reflects heterogeneous failure modes. DeepSeek-R1 shows $-$14.3 pp. Gemma 3 1B shows +12.6 pp (base-rate artefact at 60\% accuracy). Blanket-confidence models show near-zero gain. The opposite signs inflate within-group variance.

\subsection{Risk-coverage curves}

\begin{figure}[htbp!]
\centering
\includegraphics[width=0.95\columnwidth]{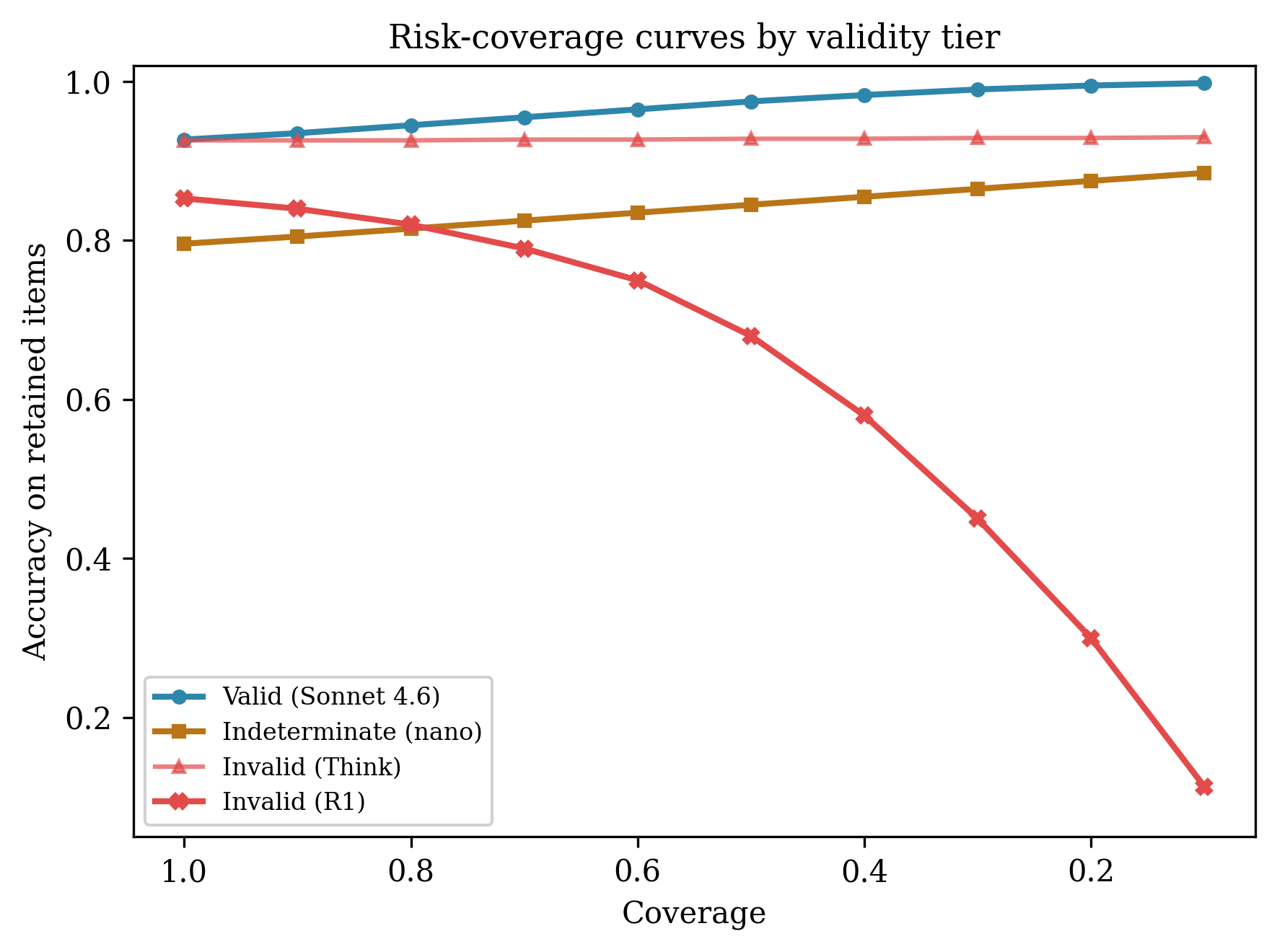}
\caption{Risk-coverage curves by validity tier. R1 shows catastrophic inversion.}
\label{fig:rccurves}
\end{figure}

Valid models show the expected pattern. Accuracy increases monotonically as coverage decreases. Sonnet 4.6 rises from 92.7\% at full coverage to 98.7\% at 30\%. GPT-5.4 rises from 92.4\% to 96.8\%.

DeepSeek-R1 shows catastrophic inversion. Accuracy drops from 85.3\% to 55.1\% at 30\% and 11.3\% at 10\%. Selecting the items R1 is most confident about selects for errors.

Gemini 3.1 Pro and Qwen 80B Think show flat curves. Neither model's confidence provides useful discrimination. Both were classified Invalid by blanket confidence (L = .967 and .974).

Gemma 3 1B (Indeterminate) shows apparent gain from 60.3\% to 81.1\% at 10\% coverage. This is a base-rate artefact. The RBS confidence interval includes zero.

\subsection{Tier classification as predictor}

One-way ANOVA on AUROC by tier. F(2,17) = 7.54, p = .005. $\eta$$^2$ = .470. The three-tier classification accounts for 47.0\% of the variance in selective prediction discrimination.

\subsection{Split-half cross-validation}

To address the same-sample limitation, we perform 1,000 random 50/50 item splits. For each split, we run the validity screen on the first half and compute AUROC on the second half. This tests whether classifications derived from one set of items predict performance on unseen items.

Median d (Valid vs Invalid) = 1.77 (95\% CI [0.95, 3.00]). P(d > 0) = 1.0 across all 1,000 splits. The effect attenuates from d = 2.81 (full sample) to d = 1.77 (split-half), as expected from reduced item counts and noisier tier assignments. It never reverses.

\subsection{Family clustering}

The 20 models cluster into seven families. The three Invalid models come from three separate families (DeepSeek, Google, Qwen). The Invalid classification is not driven by a single family. All six family representatives with sufficient data are classified Valid, confirming the Valid tier generalises across families.

\subsection{Cross-domain stability}

\textbf{Table 2.} Type 2 AUROC by cognitive track.

\begin{table}[htbp]
\centering\footnotesize
\begin{tabular}{llllllllll}
\toprule
Model & Tier & Learn & Meta & Social & Attn & Exec & Prosp & Mean & SD \\
\midrule
Sonnet 4.6 & Valid & .671 & .673 & .945 & .741 & .965 & .485 & .747 & .183 \\
Qwen Coder 480B & Valid & .603 & .558 & .713 & .850 & .791 & .691 & .701 & .110 \\
GPT-5.4 & Valid & .540 & .744 & .814 & .763 & .577 & .574 & .669 & .118 \\
Claude Haiku 4.5 & Valid & .649 & .747 & .804 & .466 & .626 & .497 & .631 & .133 \\
Gemma 3 12B & Indet. & .669 & .588 & .559 & .685 & .567 & .626 & .616 & .053 \\
Gemma 3 1B & Indet. & .506 & .588 & .553 & .542 & .431 & .514 & .522 & .054 \\
Gemini 3.1 Pro & Invalid & .428 & --- & .521 & .611 & --- & .492 & .513 & .076 \\
DeepSeek-R1 & Invalid & --- & .504 & --- & --- & --- & .000 & .252 & .356 \\
\bottomrule
\end{tabular}
\end{table}

\textit{Learn = T1 learning/overhypothesis, Meta = T2 metacognition, Social = T3 social cognition, Attn = T4 attention, Exec = T5 executive function, Prosp = T6 prospective regulation. Track numbering follows Cacioli (2026c).}

Valid models are above chance on aggregate and often above chance within tracks. But several show near-chance or below-chance performance in specific domains. Sonnet 4.6 shows .485 on Prospective. Claude Haiku 4.5 shows .466 on Attention. Every Valid model has at least one track below .55. Aggregate validity does not guarantee domain-specific validity.

Per-track r(confidence, correct) predicts per-track AUROC across all 107 model-track observations. Spearman $\rho$ = .788, p < .001. Within Valid models only, $\rho$ = .761, p < .001, n = 85. The screening construct operates at the domain level, not only at the aggregate level.

Indeterminate models show low cross-domain variability (SD = .053-.054). Their confidence signals are weak but consistent, not erratic.

\subsection{Failure modes}

Three Invalid models fail for three distinct reasons.

\textbf{DeepSeek-R1. Catastrophic inversion.} KEEPs 18.1\% of items. Those items are 25.3\% correct. The items it WITHDRAWs are 98.6\% correct. r = $-$.798. AUROC = .031. At 10\% coverage, accuracy drops to 11.3\%. A system trusting this model's confidence for abstention would selectively present errors.

\textbf{Gemini 3.1 Pro. Blanket confidence.} KEEPs 96.6\% of items. KEEP accuracy (.943) equals WITHDRAW accuracy (.944). r = $-$.001. AUROC = .522. Selective gain at all thresholds is approximately zero.

\textbf{Qwen 80B Think. Blanket confidence.} KEEPs 99.0\% of items. Five WITHDRAWs in 524 items. L = .974, r = .047 (n.s.). AUROC = .518.

\textbf{Gemma 3 1B (Indeterminate). Ambiguous signal.} Baseline accuracy 60.3\%. RBS = +.028, CI includes zero. r = $-$.031 (n.s.). AUROC = .483. Positive gain at aggressive coverage is a base-rate artefact. The Indeterminate classification is correct. The evidence for inversion is insufficient for an Invalid flag. The signal is not reliably positive either.

\subsection{Within-tier analysis}

\begin{figure}[htbp!]
\centering
\includegraphics[width=0.8\columnwidth]{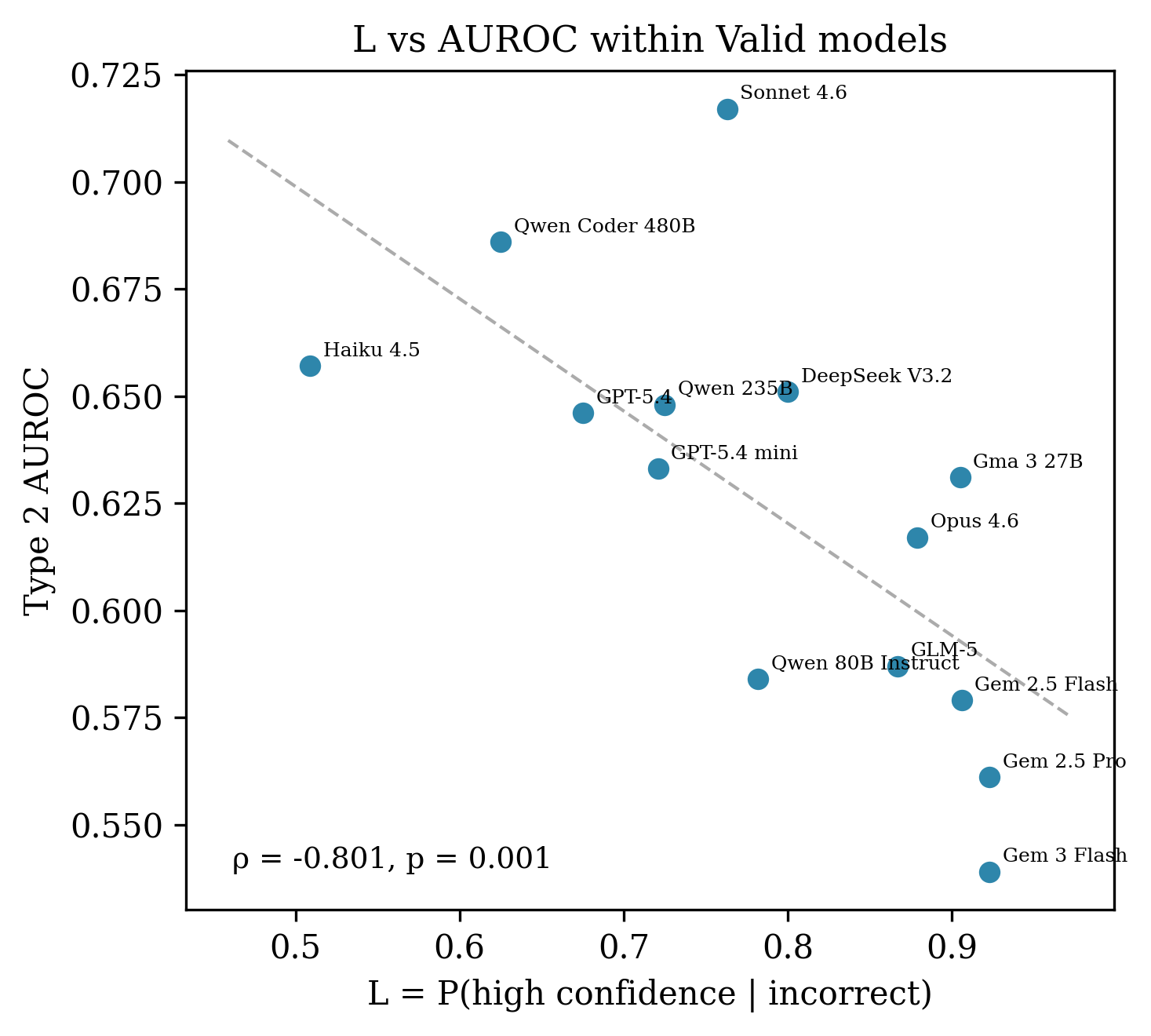}
\caption{L vs Type 2 AUROC within Valid models. Higher L correlates with lower AUROC.}
\label{fig:lauroc}
\end{figure}

Among Valid models, L (KEEP rate) correlates negatively with AUROC ($\rho$ = $-$.801, p = .001) and with selective gain at 70\% ($\rho$ = $-$.735, p = .003). r(confidence, correct) does not predict AUROC within the Valid tier ($\rho$ = +.068, p = .817).

This is a measurement ceiling. L measures the proportion of items assigned KEEP. Models with L > .90 keep nearly everything. With a 4-level ordinal scale, that leaves little variance for AUROC to exploit. Lower-L models use WITHDRAW more freely, creating more variance and higher AUROC. Once a model passes the screen, selective prediction performance is driven by probe resolution, not validity.

\section{Discussion}

\subsection{The screen predicts the criterion}

The three-tier classification predicts selective prediction performance. Valid models show above-chance AUROC, positive selective gain, and monotonically increasing risk-coverage curves. Invalid models show chance-level or inverted AUROC, zero or negative gain, and flat or decreasing curves.

The effect is large (d = 2.81, $\eta$$^2$ = .470) and robust to split-half cross-validation (median d = 1.77, P(d > 0) = 1.0).

\subsection{Heterogeneous failure, homogeneous success}

Valid models cluster tightly on AUROC (SD = .048). Invalid models show wide dispersion (SD = .231). There are many ways for confidence to be uninformative. There is essentially one way for it to be informative.

The validity indices provide mechanistic diagnosis. DeepSeek-R1 fails on Fp and RBS. Gemini 3.1 Pro and Qwen 80B Think fail on L. A system that detects "low AUROC" without understanding the failure mode cannot diagnose the problem. The indices can.

\subsection{The Indeterminate tier}

The Indeterminate tier captures models where the evidence is insufficient for a definitive classification.

GPT-5.4 nano and Gemma 3 12B show weak but positive discrimination (AUROC .565 and .615). Their confidence carries some item-level information. The signals are weaker and noisier than Valid models. In clinical terms, these resemble profiles with elevated but sub-threshold validity scales. Interpretable with caution but not with the same confidence as a clean profile.

Gemma 3 1B is different. Its AUROC (.483) is below chance. Its r is negative. Its RBS is positive. None reach threshold. It resembles a profile where the clinician suspects invalidity but cannot confirm it. The Indeterminate classification absorbs this ambiguity rather than forcing a premature binary.

For deployment, Indeterminate models should be treated as unknowns. The confidence signal may carry information. The evidence is not strong enough to rely on for safety-critical abstention.

\subsection{Cell counts and high-accuracy models}

Seven models produce fewer than 5 items in the d cell (incorrect + WITHDRAW). The screen flags these as "Insufficient data." For models with accuracy above 95\%, the screen may require 1,000+ items for stable classification. This is a practical limitation of binary probes on near-ceiling-accuracy models.

\subsection{The Gemma 3 1B reclassification}

The derivation study classified Gemma 3 1B as Invalid ("marginally inverted, borderline"). The screen classifies it as Indeterminate. RBS CI includes zero. Moving it to Indeterminate increases d from 2.09 to 2.81 and produces perfect monotonic ordering. The screen's formal rules outperform the judgement call because they respect uncertainty.

\subsection{Limitations}

\textbf{Concurrent validation.} The metrics and classifications are computed on the same dataset. This is concurrent criterion validation, not prospective generalisation. The split-half analysis partially mitigates (median d = 1.77) but does not eliminate the concern.

\textbf{Near-neighbour criterion.} The screen and the criterion both ask whether confidence discriminates correctness. The validation is conceptually proximal.

\textbf{Non-independent models.} Twenty models from seven families. Alignment can compress or distort verbal confidence (Wang et al., 2026; Wei et al., 2026). The Invalid classification is not family-specific, but the sample is too small for formal mixed-effects modelling.

\textbf{Single benchmark.} Whether the screen predicts selective prediction on MMLU, GSM8K, or ARC is untested.

\textbf{Elicitation sensitivity.} Dai (2026) shows probe design changes measured metacognitive efficiency. The results are specific to binary KEEP/WITHDRAW and BET/NO BET probes.

\subsection{Practical implications}

Screen before you deploy. Any system using LLM confidence for abstention should first verify that the confidence signal carries item-level information. The validity screen is a cheap, interpretable check.

Report the VRS Table alongside selective prediction metrics. Without it, a reported AUROC of .52 is ambiguous.

Be cautious with high-L models. Binary probes leave little room for selective prediction to exploit when the model keeps nearly everything.

\section{Conclusion}

This paper provides the first concurrent deployment-facing criterion validation of a psychometric validity screen for LLM confidence signals. The screen predicts selective prediction performance. Valid models show positive gain. Invalid models show flat or catastrophically inverted curves. The effect is robust to split-half cross-validation. $\eta$$^2$ = .470.

The validation has clear limitations. It is concurrent. It uses one benchmark. It covers one operational endpoint. Recent work on confidence faithfulness (Wang et al., 2026), scale design (Dai, 2026), and the gap between response and capability calibration (Wei et al., 2026) suggests that the present findings validate the screen against selective prediction specifically, not confidence quality in every sense.

Screen before you interpret. For selective prediction, the screen matters.

\section{Open science}

All 10,480 observations, analysis code (\texttt{validity\_screen.py} and \texttt{selective\_prediction\_analysis.py}), figures, and this manuscript are publicly available at \url{https://github.com/synthiumjp/validity-scaling-llm}. The battery is described in Cacioli (2026c), the derivation study in Cacioli (2026d), and the screening protocol in Cacioli (2026e).

\section{Generative AI}

Claude (Anthropic) was used for analysis pipeline design, code generation, and manuscript preparation. All scientific decisions, analytical design choices, and interpretive conclusions were made by the author.

\section{References}

\subsection{Clinical assessment methodology}
\begin{itemize}[nosep]
  \item Ben-Porath, Y. S., \& Tellegen, A. (2020). MMPI-3 manual for administration, scoring, and interpretation. University of Minnesota Press.
  \item Morey, L. C. (1991). Personality Assessment Inventory professional manual. PAR.
  \item Morey, L. C. (2007). Personality Assessment Inventory professional manual (2nd ed.). PAR.
  \item Rust, J., Kosinski, M., \& Stillwell, D. (2021). Modern Psychometrics: The Science of Psychological Assessment (4th ed.). Routledge.

\end{itemize}

\subsection{Selective prediction, calibration, and confidence}
\begin{itemize}[nosep]
  \item Dai, Y. (2026). Rescaling confidence: What scale design reveals about LLM metacognition. arXiv:2603.09309.
  \item Feng, S., Wan, H., Gunasekara, C., Patel, S., Joshi, S., \& Lastras, L. (2024). Don't hallucinate, abstain: Identifying LLM knowledge gaps via multi-LLM collaboration. ACL.
  \item Geifman, Y., \& El-Yaniv, R. (2017). Selective classification for deep neural networks. NeurIPS.
  \item Phillips, C., et al. (2026). Entropy alone is insufficient for safe selective prediction in LLMs. arXiv:2603.21172.
  \item Wang, Y., et al. (2026). Are LLM decisions faithful to verbal confidence? arXiv:2601.07767.
  \item Wei, J., et al. (2026). On calibration of large language models: From response to capability. arXiv:2602.13540.
  \item Wen, B., Bansal, H., Semnani, S. J., \& Lam, M. S. (2025). Know your limits: A survey of abstention in large language models. TACL, 13, 529-556.

\end{itemize}

\subsection{SDT and metacognition methodology}
\begin{itemize}[nosep]
  \item Fleming, S. M., \& Lau, H. C. (2014). How to measure metacognition. Frontiers in Human Neuroscience, 8, 443.

\end{itemize}

\subsection{Own programme}
\begin{itemize}[nosep]
  \item Cacioli, J. P. (2026a). LLMs as signal detectors. arXiv:2603.14893.
  \item Cacioli, J. P. (2026b). Do LLMs know what they know? arXiv:2603.25112.
  \item Cacioli, J. P. (2026c). The Metacognitive Monitoring Battery: A Cross-Domain Benchmark for LLM Self-Monitoring. arXiv:2604.15702.
  \item Cacioli, J. P. (2026d). Before you interpret the profile: Validity scaling for LLM metacognitive self-report. arXiv [companion paper].
  \item Cacioli, J. P. (2026e). Screen before you interpret: A portable validity protocol for benchmark-based LLM confidence signals. arXiv [companion paper].

\end{itemize}


\end{document}